\documentclass[conference]{IEEEtran}
\IEEEoverridecommandlockouts
\usepackage{cite}
\usepackage{amsmath,amssymb,amsfonts}
\usepackage{algorithmic}
\usepackage{graphicx}
\usepackage{textcomp}
\usepackage{xcolor}
\usepackage{enumitem}
\def\BibTeX{{\rm B\kern-.05em{\sc i\kern-.025em b}\kern-.08em
    T\kern-.1667em\lower.7ex\hbox{E}\kern-.125emX}}
\begin{document}

\title{Petroleum prices prediction using data mining techniques – A Review \\
}

\author{\IEEEauthorblockN{John Ngechu}
\IEEEauthorblockA{\textit{dept.Computer Science } \\
\textit{Dedan Kimathi University of Technology}\\
johnngechu18@gmail.com}
\and
\IEEEauthorblockN{Kiplang'at Weldon}
\IEEEauthorblockA{\textit{dept. Computer Science} \\
\textit{Dedan Kimathi University of Technology}\\
welldonekplc@gmail.com}
\and
\IEEEauthorblockN{Ngatho Everlyne}
\IEEEauthorblockA{\textit{dept. Computer Science} \\
\textit{Dedan Kimathi University of Technology}\\
ngathoevedkut@gmail.com}
\and
\IEEEauthorblockN{Nancy Njambi}
\IEEEauthorblockA{\textit{dept. Computer Science} \\
\textit{Dedan Kimathi University of Technology}\\
nancynjambi2016@gmail.com}
\and
\IEEEauthorblockN{Kinyua Gikunda}
\IEEEauthorblockA{\textit{dept. Computer Science} \\
\textit{Dedan Kimathi University of Technology}\\
patrick.gikunda@dkut.ac.ke}
}

\maketitle

\begin{abstract}
Over the past 20 years, Kenya's demand for petroleum products has proliferated. This is mainly because this particular commodity is used in many sectors of the country’s economy. Exchange rates are impacted by constantly shifting prices, which also impact Kenya's industrial output of commodities. The cost of other items produced and even the expansion of the economy is significantly impacted by any change in the price of petroleum products. Therefore, accurate petroleum price forecasting is critical for devising policies that are suitable to curb fuel-related shocks. Data mining techniques are the tools used to find valuable patterns in data. Data mining techniques used in petroleum price prediction, including artificial neural networks (ANNs), support vector machines (SVMs), and intelligent optimization techniques like the genetic algorithm (GA), have grown increasingly popular. This study provides a comprehensive review of the existing data mining techniques for making predictions on petroleum prices. The data mining techniques are classified into regression models, deep neural network models, fuzzy sets and logic, and hybrid models. A detailed discussion of how these models are developed and the accuracy of the models is provided.
\end{abstract}


\section{Introduction}
The most important source of energy for economic activity is petroleum, which is also the most traded commodity on the globe\cite{b1}. Petroleum is essential to industry and civilization, and as it supplies a sizable amount of the world's energy needs, it plays a significant role in world politics and international relations\cite{b2}. With 33\% of the primary energy consumed globally, it continues to be the most significant energy source in the world. Its dominance in the transportation industry, which accounts for 94\% of all energy used, highlights its importance\cite{b3}.The expense of fuel has long caused concern across all social groups. Fluctuations in the price of petroleum have a direct impact on the economy of the nation, impacting retailers, consumers, the production industries, the public and private transportation sectors. Additionally, it has an indirect effect on the price of commodities, the stock market, inflation, and other significant aspects of modern man's life cycle\cite{b4}. 

Numerous variables affect the price of petroleum. First, just as with any other commodity, economic factors such as petroleum supply and demand greatly influence the rise and fall of petroleum prices\cite{b5}. Second, the price of petroleum is influenced by the chemical characteristics of oil, such as density and viscosity, as well as by technical elements of the oil sector, such as refining capacity and above-ground oil stocks. Additionally, production techniques like shale oil extraction have a substantial influence on the supply of crude oil and, consequently, the price of petroleum\cite{b6}. Third, geopolitical conflicts and tensions have a big impact on oil prices. Because the Middle East has the largest oil reserves in the world and is a politically unstable region, geopolitical factors like the Russia-Ukraine war have had a big influence on oil prices\cite{b7}. By impacting the supply of petroleum, economic sanctions imposed on significant oil-producing nations can also cause disruptions in the world energy markets.

To shed light on the behavior of oil prices, several attempts have been made. Two key categories are taken into consideration while estimating the price of petroleum. The first group includes conventional statistical and econometric approaches.This includes the exponential smoothing model (ESM), linear regression (LinR), autoregressive integrated moving average (ARIMA), generalized autoregressive conditional heteroskedasticity (GARCH), random walk (RW), and error correction models (ECM)\cite{b8}. Only linear trends may often be detected in time-series data using traditional statistical and econometric tools. These models may fall short of capturing the nonlinear characteristics of crude oil pricing\cite{b9}. To get around this constraint, Data Mining techniques using AI models\cite{b10}with strong self-learning abilities are used.

Data mining is the process of discovering patterns and trends to extract valuable information from large data collections to assess or make a decision\cite{b11}. In order to find the valuable patterns in data, data mining techniques are the tools used\cite{b12}. Data mining techniques used in petroleum price prediction includes, artificial neural networks (ANNs), support vector machines (SVMs), and intelligent optimization techniques like the genetic algorithm (GA) have grown increasingly popular\cite{b13}.

Due to a multitude of natural factors, petroleum prices are very erratic, vary more than other mineral resources, and are challenging to predict\cite{b3}. On the one hand, accurate forecasting is critical for private investors and central banks, who rely on accurate oil price forecasts to devise suitable policies in reaction to petroleum-related shocks. As a result, a wide spectrum of organizations seeks precise oil price forecasts\cite{b10}. 

Since petroleum is the most major source of energy and is used in practically every area of our life, it has a big impact on how our economy develops.As a result, shifts in the price of petroleum are frequently excellent indications of shifts in the national economy and worldwide markets\cite{b14}. The ability to anticipate these swings in petroleum prices helps economic players, such as companies, to adapt to impending market changes and provides decision-makers with accurate knowledge with which they may use it to decide on the best course of action\cite{b15}.

The following are some reasons for doing the research: a) Data mining techniques help quickly automate predictions of behaviors and trends and discover hidden patterns. (b) Data mining has been used in many domains for price prediction. (c) Data mining helps ease the process of analyzing enormous amounts of data. (d) Data mining helps in making informed decisions. This study aims to discover petroleum price prediction research and development and provide insight into the implementation of data mining techniques in relation to petroleum price prediction. As a result, the investigation's main focus will be on how well data mining tools can forecast petroleum prices in order to inform future researchers. In this light, the study’s research questions are: a) What is the role of data mining in petroleum price prediction? b) What type of data mining techniques should be used? c) What are the benefits of using data mining techniques in petroleum price prediction in Kenya?

\section{METHODOLOGY}
In order to address the research questions, a bibliographic analysis of the subject under study was carried out between 2019 and 2022. There were two steps involved: (a) clustering of related works and (b) detailed review and analysis of the works. In the first stage, a keyword-based search is conducted, utilizing all possible combinations of two sets of keywords, the first of which is concerned with data mining techniques and the second group refers to petroleum price prediction (i.e., crude oil price, oil price, fuel price). The analysis was done while considering the following research questions: (a) petroleum product predicted, (b) data set used, (c) based on the author’s performance metric, and accuracy, and (d) data mining techniques used.

\section{AN OVERVIEW OF DATA MINING TECHNIQUES}
Data mining is the process of detecting or extracting fascinating patterns, correlations, changes, anomalies, and key structures from enormous amounts of data held in various data sources such as file systems, databases, data warehouses, and other information repositories\cite{b16}. As a multidisciplinary field, data mining utilizes a number of techniques or approaches from various disciplines, including statistics, machine learning, neural networks, genetic algorithms, fuzzy sets, and visualization \cite{b17}.

\subsection{Single Statistical and Econometric Prediction Model}
Regression models are the fundamental time series forecasting methods. The ARIMA equation, a linear equation, is the most often used regression model\cite{b18}. ARIMA model has been widely used to predict fuel prices because of its ease of use\cite{b19}. The ARIMA (p,d,q) standard model is described as:
\begin{equation}
\varphi_p (B)\nabla^d y_t=\mu+\theta_q (B)\varepsilon_t
\end{equation}
\cite{b21} applied the autoregressive integrated moving average model to predict Fuel prices in Malaysia. The data source that was used for that particular study was the weekly price of Ron97 which was made public by the ministry of trade in Malaysia. ARIMA model used to forecast petroleum fuel prices in Malaysia has been proven to predict the fuel prices accurately for shorter periods as compared to the longer periods. A study was conducted by \cite{b22} to improve the predictions of fuel cost distribution using the ARIMA model.The data for this study was obtained from the US Energy Information Administration (EIA)-923 from January 2013 to December 2016. Data selection is done first in order to explore the data and come up with a fuel cost time series. The fuel price time series is in turn used in the ARIMA model the fuel price for the next month. To validate the model, 6 months are used from July 2016 to December 2016. A normal distribution is then used to fit the data to obtain an estimated fuel price. Kullback-Leibler divergence is then used to evaluate the performance of the algorithm \cite{b23}. The results from the predictions indicate that the model have higher accuracy as compared to using three-month delayed EIA data when the market does not have high volatility.   
\cite{b24} proposed a model to predict petroleum prices in Kenya using the Autoregressive integrated moving average (ARIMA) and the Vector Autoregressive models(VAR). The data sources from the research were crude oil prices in the world market, exchange rates, inflation rates and the secondary petroleum pump prices data from the Energy and Petroleum Regulatory Authority of Kenya. The data used was from January 2011 to December 2018. After developing the model’s prediction was made for the next twelve months. The petroleum prices were discovered to be non-stationery. The prediction accuracy of the two models ARIMA and VAR was compared using root mean square error (RMSE), mean absolute error (MAE) and mean absolute percentage error (MAPE).

\begin{table}[htp]
    \centering
    \caption{Comparison of ARIMA and VAR model\cite{b24}.}
    \includegraphics[width=8cm]{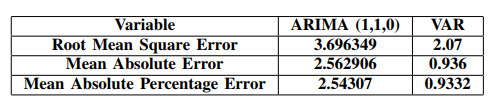}
    \label{tab:Tab 1:}
\end{table}

From the study, VAR was better in the prediction of the petroleum prices in Kenya as compared to the ARIMA model.
\subsection{Deep Neural Networks.}

Deep learning is a subset of machine learning techniques that is based on learning data representations\cite{b25}.Data extraction and manipulation may be carried out automatically using neurons, a network of multi-layered non-linear processing units utilized in deep learning models. An artificial neural network is a network made up of these neurons. Node clusters in ANN layers are joined to produce a picture of the human brain. The input for the subsequent layer is the yield of the previous layer. These system models typically pick up new information from their training data \cite{b26}
\begin{enumerate}[label=(\alph*)]
\item CNN: A novel crude oil price forecasting approach based on online media text mining to capture the more immediate market antecedents of price swings\cite{b27}. Based on the CNN model, sentiment analysis, and topic identification (Latent Dirichlet Allocation (LDA) topic model), a text-based topic-sentiment synthesis technique was presented for time series generation. By taking into account unanticipated political or social developments that are reported in the press, this strategy could improve crude oil price prediction accuracy. The CNN model's accuracy was 61 percent, which was lower than projected. Rolling test windows for each and every news headline, like in normal time series forecasting, yielded a lower accuracy of 59.8

Selvin, S. used CNN, LSTM, and RNN architectures to forecast the price of NSE-listed companies and compared their performance. The final results showed that CNN is the best architecture to predict the price of stock because it could identify the trend of the directional change \cite{b28}. Liu, S. proposed a CNN-LSTM model, and the model performed a basic momentum strategy and benchmark model for which the return rates were 0.882 and 1.136, respectively. The CNN part could extract useful features even from low signal-to-noise time-series data, and the LSTM part could predict future stock prices with high accuracy. Then, the predicting outcomes were used as timing signals \cite{b29}.

\item LSTM: Long Short-Term Memory (LSTM), also known as an improved recurrent neural network (RNN) model. In RNN, the output from the previous step is provided as input during the current phase. It addressed the issue of long-term dependency of RNNs, in which the RNN can provide more precise forecasts based on current data but cannot anticipate the word hold on in the long-term memory. By default, LSTM will hold onto the information for a long time. On the basis of time-series data, it is utilized for processing, predicting, and categorizing \cite{b30}.

The issue with stock market forecasting is that it relies heavily on historical data, huge numbers, and a significant volume of data. As a result, LSTM \cite{b31}helps RNNs by storing data for subsequent usage, which helps to control error. As the stage goes on, the forecast gets more accurate proving that it is far more reliable than other methods. To speed up training and prevent overfitting,\cite{b32} suggested an LSTM model with two stacked LSTM layers, an output value of 256, and a dropout ratio of 0.3. The stacked LSTM model received a test score of 0.00875 MSE.

To enhance the prediction, \cite{b33} has taken use of the relationship between stock price and sentiment. Use of live stock prices from the Yahoo Finance API, web headlines, historical stock data, and sentiments from Twitter. The univariate and multivariate LSTM were both fed sentiment scores from twitter, the web, and Google Trend. Through the use of Weighted Average Ensemble, the models were combined. As the days passed, the accuracy fell. The highest prediction score accuracy was 0.99 for the first day, 0.92 for seven days, and thereafter 0.62 until the thirty-first day.

\item RNN: RNNs are a type of neural network that are constructed to handle sequential input \cite{b26}. They are composed of regular recurrent cells, in which the networks between the neurons form a directed graph, or, to put it another way, the hidden layers contain a self-loop. This enables RNNs to use both the information they have acquired in the past and the present input to learn the current state by utilizing the hidden neurons’ previous state \cite{b31}. They can pick up a variety of skills because of this. They include things like speech and handwriting recognition. Chen, W. proposed an RNN-Boost model that made use of the technical indicators and sentiment features, and Latent Dirichlet allocation (LDA) features to predict the price of stocks. Its results showed that the proposed model outperformed the single-RNN model  \cite{b34}.
RNN was used \cite{b4} to create a model to forecast fuel costs in Chennai, Kolkata, Mumbai, and Delhi for the next day. The past prices in these cities are kept up to date using the international Brent oil barrel price and currency rates, and the next day's gasoline costs are forecasted. Based on their observations, they concluded that the anticipated values are quite similar to the real prices. For next-day pricing, the model's accuracy is higher than 90
\end{enumerate}

\subsection{Hybrid Models}
Recent research has developed a method for mining vast volumes of data to draw out useful information using a combination of several data mining techniques. This section is a review of hybrid techniques that have been used to predict petroleum prices. Hybrid models have a better prediction accuracy and are more robust since they can combine the individual advantages of the techniques that are used.

In Lake’s study, time-series data mining techniques were used to construct a model for projecting retail oil and natural gas vehicle costs for vehicles in Thailand. All three linear regressions, multi-layer perceptron, and support vector machines for regression were employed  \cite{b34}. The experiment's time series was separated into six data sets by the researcher. Gasohol 91, Gasohol 95, Gasohol E20, Gasohol E85, and Ultra Force Diesel and natural gas for vehicles (NGV), with average prices of 30.7835-baht, 31.9928-baht, 28.6693-baht, 21.4488-baht, 27.4076-baht, and 12.1327-baht, respectively.

There were two parts to the research: a) For assessing the model's efficiency, use Root Mean Square Error (RMSE) and Mean Absolute Error (MAE). b) evaluating the model's performance using the Magnitude of Relative Error (MRE), which was separated into monthly tests. The following were the outcomes: a) For gasoline E85 and ultra-high-pressure diesel, linear regression showed the best fit, with MMRE rates of 2.46 and 4.60, respectively. b) For gasohol 91, gasohol 95, gasohol E20, and NGV, which had MMRE rates of 3.69, 3.20, 3.54, and 6.89, respectively, the support vector machine for regression was the most appropriate.

A hybrid model that combines data mining techniques such as nonlinear autoregressive (NAR) neural network, autoregressive integrated moving average model (ARIMA)and exponential smoothing model (ESM) was implemented in a study done by \cite{b9}. These data mining techniques were combined in a state-space model framework with the aim of increasing the accuracy of prediction of prices. The three models, ESM, ARIM and NAR are combined in a linear form. The next step is to determine the weights of the hybrid model using three approaches; use of constant equal weights, use of constant genetic algorithm weights and use of time varying weights\cite{b36}. The model is finally evaluated using mean absolute error (MAE), root mean-square error (RMSE), mean absolute percentage error (MAPE) and directional accuracy. From the study the hybrid model was able to outperform its constructive models.

\subsection{Fuzzy Set and Logic }
Instead of the traditional ”true or untrue” (1 or 0) Boolean operators on which the contemporary computer is built, fuzzy logic bases computation on ”degrees of truth.” In the 1960s, Lotfi Zadeh of the University of California, Barkley, was the first to put forth the concept of fuzzy logic \cite{b37}. On the other hand, a class of items with a range of membership grades is
referred to as a fuzzy set. A membership (character) algorithm that awards each object a membership grade ranging from zero to one defines such a set. Fuzzy logic and fuzzy set are widely applied in counterparty society in the prediction of commodities such as petroleum and gases by oil exporting companies and manufacturers \cite{b8}.

\begin{enumerate}[label=(\alph*)]
\item OPEC Oil Price Prediction Using ANFIS: : In this study, the Organization of Petroleum Exporting Countries (OPEC)’s oil prices are forecasted using an adaptive neuro-fuzzy inference system (ANFIS) (OPEC). The suggested feature set given to the ANFIS is the innovative part of the proposed model \cite{b37}. The proposed approach is examined in the numerical models using an example of simulating the time series of OPEC oil. In comparison to traditional neural networks, ANFIS with the suggested variable set exhibits greater accuracy in predicting oil prices.
\item Forecasting the Direction of Short-Term Crude Oil Price Changes with Genetic-Fuzzy Information Distribution: The short-term oil price forecasting challenge is addressed in this research using a unique approach \cite{b38}. This method uses a new fusion genetic-fuzzy information sharing method to perform the short-term oil price series’ association with incomplete fuzzy information. Next, a workable coding method based on multidimensional knowledge controlling points is used to adapt the genetic-fuzzy dissemination of information to time series forecasting. The empirical study findings show that the innovative fused genetic-fuzzy information sharing approach statistically beats the standard logistic regression model in terms of prediction accuracy using the market prices of West Texas Medium (WTI) and London crude oil as sample data. The outcomes show that this new strategy is successful in terms of direction accuracy.

In a study carried out by \cite{b8}, a model predicting crude oil price behavior in the future was developed. The model included data from the previous year as well as the actual oil price from the day before, allowing for more precise forecasting. Their two-phase concept was built around fuzzy time series. In the second step, they used the largest integer function to combine the forecasts from their series and the actual crude oil prices from the previous day. Surprisingly, the model captured the chaotic and non-linear behavior of daily crude oil prices for 2021. ANN and SVM, on the other hand, are known to provide superior outcomes in crude oil price prediction than traditional approaches, thus these models were employed to statistically compare them to their model. The suggested model outperformed existing models in terms of both the RMSE and the MAE values, notably ANN and SVM.     
\end{enumerate}

\begin{table}[htp]
    \centering
    \caption{comparison of the RMSE and MAE}
    \includegraphics[width=8cm]{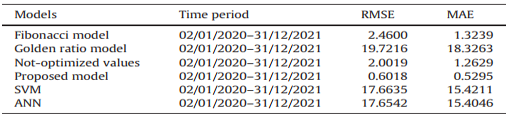}
    \label{tab:Tab 2:}
\end{table}

\section{The critiques of the existing models.}

Numerous issues can be discovered in light of the models that has been evaluated so far. First off, additional market-aligned inputs are not taken into account because the data used in the projections is often derived from the WTI or Brent price. Because of its reliance on a number of variables, the market for petroleum prices is prone to volatility\cite{b38}. A prediction model credibility might be damaged by ignoring these aspects, which prevents it from being complete \cite{b33}. A model that produces accurate predictions shows strong linkages between inputs and outputs, indicating the presence of dependency \cite{b8}. 

Second, there are still few studies that focus on the market's volatility. The majority of discussions have centered more on the price side of the forecast than on the causes of the fluctuations\cite{b7}. Although the supply and demand of oil both significantly affect price volatility, the inclusion of these facts merely restricts the potential of other elements, such as input data, making a model incomplete \cite{b5}. An accurate market prediction may be made by integrating and linking the important contributing components.

Third, time-series data have been used in the majority of the models examined \cite{b9}. In the majority of the research, there was no data pretreatment or data representation method. These two procedures aid in normalizing data sets, removing noise from them, and cleaning up data sets to assist condense the prediction process and ultimately produce reliable results\cite{b16}. The prediction model will be less reliable without these operations.

Fourth, research has revealed that forecasting price trends is more common than forecasting discrete prices\cite{b27}. Even if the applicability of the studies completed so far is still debatable, discrete price forecasts will make research more appealing and useful for practitioners. 

\section{CONCLUSION AND FUTURE WORK}

The vast amount of prior research has primarily employed quantitative analytical techniques. This is due to the availability of historical market data and recent developments in statistical techniques\cite{b27}. Most of these studies have placed a significant reliance on the publication of official macroeconomic statistics, which are gathered, examined, and aggregated by regulatory agencies\cite{b8}. Politics, natural disasters, and emergent occurrences have not previously been studied in the context of quantifiable evidence. Although it comes with more complexity, including qualitative data into petroleum predictions sounds promising. Compared to official statistics, information on the internet is updated more regularly. For locating viewpoints and obtaining information, text mining techniques are helpful. The development of more sophisticated forecasting methods is necessary to increase the accuracy of petroleum forecasting. These new data sources must be integrated with high-frequency and lag-free data\cite{b27}.

\end{document}